\documentclass[conference]{IEEEtran}

\newtheorem{problem}{Problem}
\usepackage{microtype}
\usepackage{graphicx}
\usepackage{subfigure}
\usepackage{booktabs} 
\usepackage{hyperref}
\usepackage{url}            
 
\usepackage{amsfonts}       
\usepackage{nicefrac}       

\usepackage{graphicx} 
\graphicspath{{figures/}}
\usepackage{epsfig} 
\usepackage{ragged2e}
\usepackage{amsmath}
\usepackage{amssymb}
\usepackage{epstopdf}
\usepackage{algorithm}
\usepackage{algorithmic}
\usepackage{multirow}
\usepackage{rotating}
\usepackage{subfigure}
\usepackage{color}
\usepackage{mysymbol}
\usepackage{url}

\begin{document}

\title{Real-Time Detectors for Digital and Physical Adversarial Inputs to Perception Systems}

\author{Yiannis Kantaros, Taylor Carpenter, Kaustubh Sridhar, Yahan Yang, Insup Lee, James Weimer \thanks{The authors are with the School of Engineering and Applied Sciences, University of Pennsylvania, Philadelphia, PA 19103, USA, {\tt\footnotesize \{kantaros,carptj,ksridhar,yangy96,lee,weimerj\}}{\tt\footnotesize{@seas.upenn.edu}.}}
}
\maketitle


\begin{abstract}
Deep neural network (DNN) models have proven to be vulnerable to adversarial digital and physical attacks. In this paper, we propose a novel attack- and dataset-agnostic and real-time detector for both types of adversarial inputs to DNN-based perception systems. In particular, the proposed detector relies on the observation that adversarial images are sensitive to certain label-invariant transformations. Specifically, to determine if an image has been adversarially manipulated, the proposed detector checks if the output of the target classifier on a given input image changes significantly after feeding it a transformed version of the image under investigation. Moreover, we show that the proposed detector is computationally-light both at runtime and design-time which makes it suitable for real-time applications that may also involve large-scale image domains. To highlight this, we demonstrate the efficiency of the proposed detector on ImageNet, a task that is computationally challenging for the majority of relevant defenses, and on physically attacked traffic signs that may be encountered in real-time autonomy applications. Finally, we propose the first adversarial dataset, called AdvNet that includes both clean and physical traffic sign images. Our extensive comparative experiments on the MNIST, CIFAR10, ImageNet, and AdvNet datasets show that VisionGuard outperforms existing defenses in terms of scalability and detection performance. We have also evaluated the proposed detector on field test data obtained on a moving vehicle equipped with a perception-based DNN being under attack.
\end{abstract}


\section{Introduction}\label{sec:introduction}

\begin{figure}[t]
  \centering
    \includegraphics[width=0.8\linewidth]{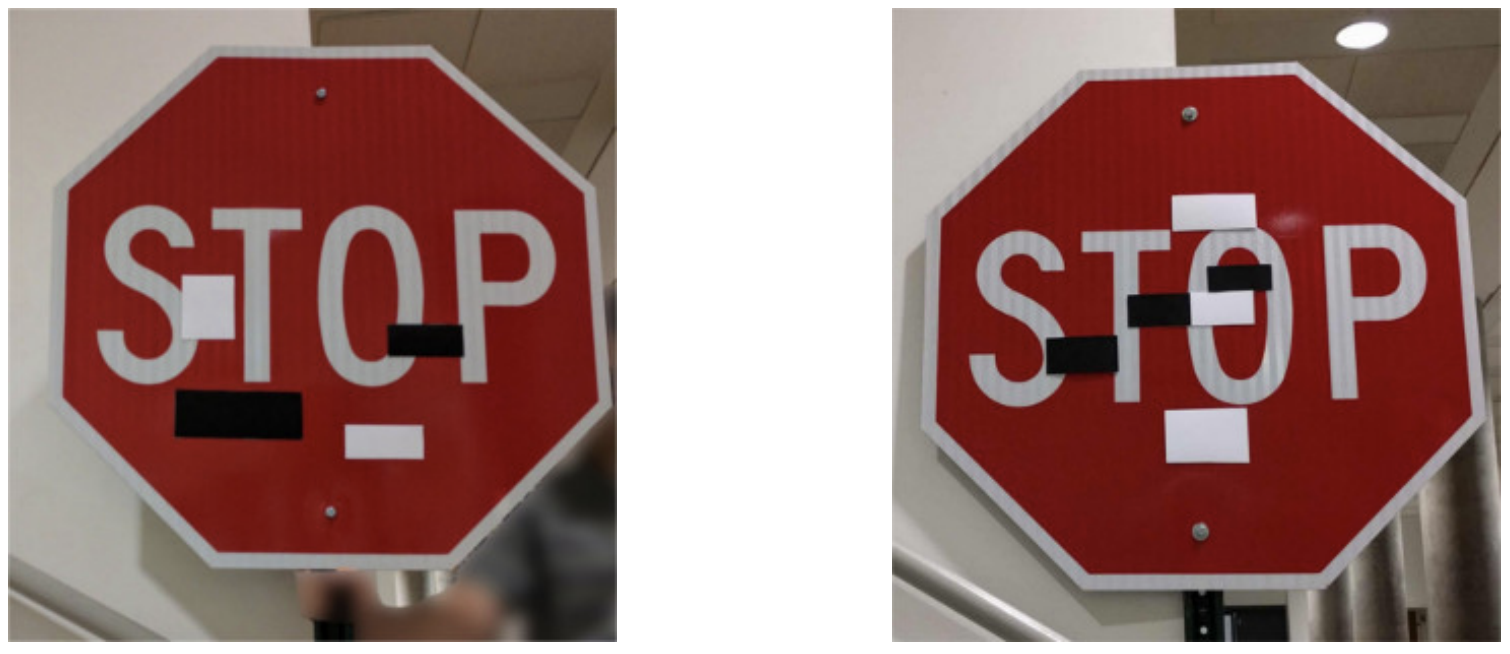}
  \caption{Examples of adversarial stickers that fool the LISA-CNN (left) \cite{eykholt2018robust} and GTSRB-CNN (right) \cite{stallkamp2012man} so that the stop sign is missclassified as speed limit 35 sign; picture borrowed from \cite{eykholt2018robust}.}
  \label{fig:attack_example_p}
\end{figure}

\begin{figure}[t]
  \centering
    \includegraphics[width=1\linewidth]{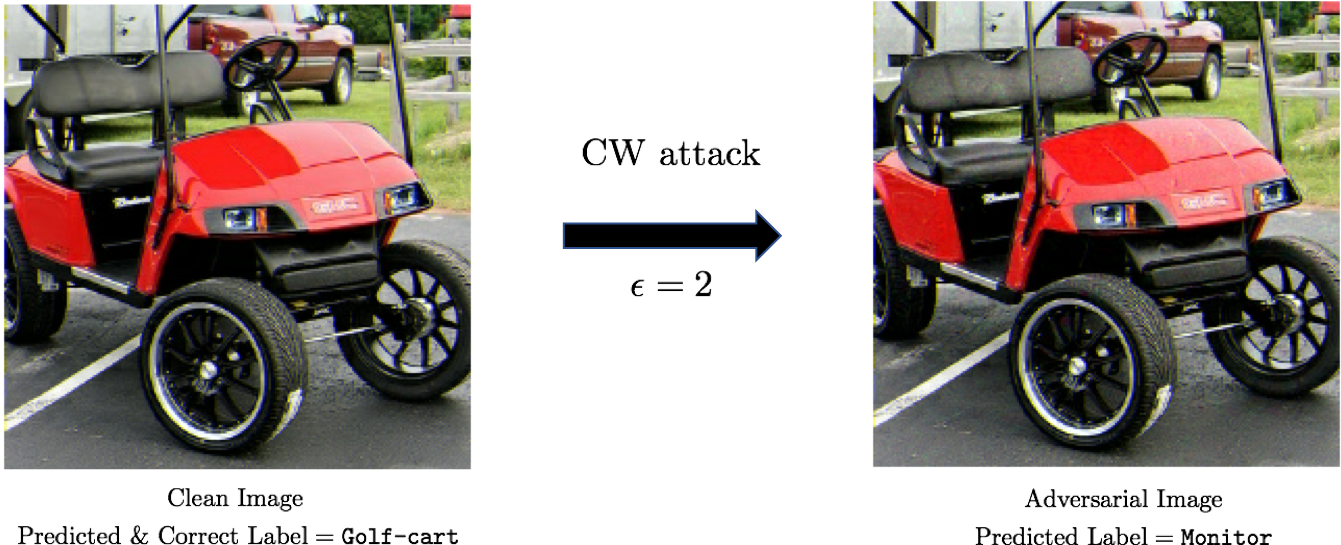}
  \caption{Examples of almost imperceptible adversarial images from ImageNet after CW attack.}
  \label{fig:attack_example_d}
\end{figure}

Deep neural networks (DNNs) have been deployed in multiple safety-critical systems, such as medical imaging, autonomous
cars, and surveillance systems. At the
same time, DNNs have been shown to be vulnerable to adversarial
examples~\cite{szegedy2013intriguing}, i.e., inputs which have deliberately been modified to cause either misclassification or desired incorrect prediction that would benefit an attacker. Adversarial examples in the literature can be divided into two sub-classes depending on how the attack is executed.  One augments the physical environment to induce misclassification (e.g., adding a sticker to a stop sign) \cite{karmon2018lavan,eykholt2018robust}, while the other adds a small perturbation to the classifier input data; see Figures \ref{fig:attack_example_p}-\ref{fig:attack_example_d}. 

Adversarial examples, especially in the case of image classification that is also considered in this paper, have received increased research attention due to the following properties. First, the difference between legitimate and adversarial digital inputs can be imperceptible, making adversarial detection a very challenging task~\cite{szegedy2013intriguing}. Second,
the transferability of adversarial digital samples between different models allows for black-box attacks~\cite{goodfellow2014explaining,zantedeschi2017efficient}. Third, the robustness of physical attacks against various environmental conditions and backgrounds \cite{eykholt2018robust}. Fourth, both digital and physical adversarial samples are often misclassified with high confidence, implying that DNNs fail to discriminate between adversarial and legitimate inputs~\cite{guo2017calibration,eykholt2018robust}.

%

To establish reliability and security of DNN-based perception systems against adversarial input images, we propose, VisionGuard (VG), a novel attack- and dataset-agnostic detection framework. VG does not modify the specific classifier and, does not rely on building separate classifiers. 
Instead, VG relies on the observation that adversaries may be successful at fooling DNNs due to the large feature space over which they can look for adversarial inputs. This is also validated in our experiments: the larger the input space (i.e., image dimensions), the easier to fool the target classifier. Motivated by this, the proposed defense aims to shrink the feature space available to adversaries. In particular, to determine if an image is adversarial, VG checks if the softmax output of the target classifier on a given input image changes significantly after feeding it a refined version of that image. To refine images, i.e., to squeeze out possibly unnecessary features of the input image, we employ label-invariant transformations. Specifically, our experiments suggest that lossy compression (e.g., JPEG) with high compression quality and brightness/rotation transformations can achieve high detection performance against digital and physical adversarial attacks, respectively; however, any other transformation can be integrated with the proposed detector. 
Then, we measure the similarity of the corresponding softmax outputs using the Kullback–Leibler (K-L) divergence metric. If this metric is above a threshold, the image is classified as adversarial; otherwise, it is classified as clean. 
Finally, we would like to emphasize that in this paper we are particularly interested in real-time applications (e.g., autonomous cars) imposing runtime constraints on both the attacker and defender that are typically ignored in the related literature. In particular, the attacker has to be capable of generating adversarial noise at a rate that is greater than the rate at which images are received and classified by the target perception system  while the detector has to reason about the trustworthiness of input images at a rate that is smaller than the attack rate. As a result, we focus on attacks that are either fast enough to craft at runtime (e.g., defense-oblivious digital attacks such as FGSM and CW \cite{carlini2017towards} 
or physical attacks that are generated offline but they can fool perception systems at runtime \cite{eykholt2018robust}. Observe that this precludes white-box attacks - that assume perfect knowledge of both the target DNN and the defense mechanism - due to their high computational cost. Nevertheless, such attacks do not concern current users, since it is almost impossible to get such complete information \cite{chen2019damagenet}.

\subsection{Related Works}
Similar defenses that rely on image transformations have also been proposed in the image purification domain to defend against \textit{digital} attacks only. For instance, \cite{guo2017countering,das2018shield,samangouei2018defense,yin2020defense} apply compression, bit depth reduction, crop ensemble to remove noise and possible adversarial digital components from images. Purification is applied to \textit{all} images, whether they are adversarial or not, which compromises the accuracy of the network on clean images \cite{zheng2016improving,kou2019enhancing}. In contrast, VG is orthogonal to these approaches as it focuses on \textit{detection}, and not purification, of both digital and physical attacks, without affecting the accuracy of the target classifier. 
Image transformations have also been employed in \cite{tian2018detecting} to detect adversarial inputs but in a completely different way than the proposed one. In particular, \cite{tian2018detecting} relies on building a DNN-based detector that takes as input $K\times N$ features, where $K$ is the number of applied transformations (e.g., rotation and translation) and $N$ is the number of logits/classes. For instance, for the MNIST dataset, the authors consider $K=45$ and $N=10$ while for ImageNet there are $N=1000$ logits. 
This large input space for the DNN detector may render its training computationally challenging especially for large image domains; in fact, evaluation on ImageNet is not provided in \cite{tian2018detecting}. 
%
Similar to VG, MagNet \cite{meng2017magnet} checks if an input image is adversarial by applying a single image transformation and examining the corresponding softmax output. Specifically, MagNet employs auto-encoders - as opposed to label-invariant 
transformations considered in this paper - to generate new images that are reconstructed from the original ones. 
Common in \cite{tian2018detecting,meng2017magnet} is that the proposed detectors are dataset-specific, as a new autoencoder/DNN-detector needs to built for each dataset. To the contrary, as shown by our extensive experiments VG is dataset-agnostic; see Section \ref{sec:eval} and \ref{sec:sim}. 

Adversarial detectors that do not rely on image transformations have also been proposed in  
\cite{feinman2017detecting,jha2018detecting, fidel2020explainability}. 
Particularly, in \cite{feinman2017detecting,jha2018detecting} the kernel density estimation (KDE) detector is proposed that selects thresholds on the likelihood of an image. This likelihood is computed using the outputs of the last hidden layer of the classifier for the image under investigation and for all training images.  
Note that the defense proposed in \cite{feinman2017detecting,jha2018detecting} has also been employed in \cite{pang2018towards} which, however, requires training of the target classifier using a reverse-cross entropy objective function. 
\cite{fidel2020explainability} proposes a DNN detector trained using the internal layers of the target DNN classifier to discriminate between normal and adversarial inputs. A conceptually similar defense is proposed in \cite{cohen2020detecting} that relies on training a detector using information provided by the DNN activation layers. 
Recently, detectors that operate directly on images, independent from the targeted classifier, have been proposed that rely on steganalysis methods \cite{liu2019detection}. However, the defense in \cite{liu2019detection} is attack-specific, since separate detectors must be designed for each type of attack. To the contrary VG is an attack-agnostic detector; see Section \ref{sec:eval} and \ref{sec:sim}. 


We would like to highlight that the majority of all defenses discussed above have only been evaluated on digital attacks and on smaller datasets such as MNIST and CIFAR10. Therefore, their applicability on more realistic physical attacks and on large-scale image domains is questionable as also discussed in \cite{carlini2017adversarial} and shown in our comparative experiments; see Section \ref{sec:sim}. Finally, we would like to emphasize that existing detectors  - see e.g., \cite{meng2017magnet,tian2018detecting,feinman2017detecting,jha2018detecting,pang2018towards,cai2020real} - are dataset-specific as they heavily rely on training sets directly and/or building separate DNN models based on these training sets. Particularly, in these works, a new auto-encoder, a new DNN-based detector need to be built, or new embeddings need to be extracted and stored for each dataset. Therefore, it is unclear how these methods perform when they are deployed in real-world environments for which datasets do not exist. To the contrary, VG does not rely on training sets, or on the training process of the target DNN, or on building separate DNN classifiers. For instance, we show through extensive experiments that VG is \textit{dataset-agnostic} in the sense that the same transformation (JPEG with compression quality $92\%$) yields high detection performance across digitally attacked datasets that differ both in content and in image dimensions.

An alternative research direction to design defense mechanisms - that do not rely on image transformations - focuses on improving robustness of   the target DNN via adversarial/robust training approaches. Specifically, adversarial training methods rely on augmenting the training set with adversarial examples and incorporating an adversarial component as a regularizer in the classification objective; see e.g., \cite{moosavi2016deepfool,kurakin2016adversarial} and the references therein. Additionally, distillation, initially proposed to reduce the size of deep neural networks  \cite{hinton2015distilling}, can be used as a mechanism to train roust DNNs \cite{papernot2016distillation}.  
Similar, but more computationally efficient training methods, called label smoothing, have also been proposed in \cite{warde201611}. 
A computationally-efficient robust training method is also proposed in \cite{zantedeschi2017efficient} that relies on Gaussian data augmentation during training and requires the BReLU activation function.  Note that adversarial/robust training methods \textit{complement} adversarial detectors such as VG. A recent summary of existing defenses can also be found in \cite{yuan2019adversarial}. 

\subsection{Evaluation: Scalability \& Detection Performance}\label{sec:eval}
We evaluate VG against digital attacks on the MNIST, CIFAR10, and ImageNet datasets and we show that, unlike relevant works, it is very computationally light in terms of runtime and memory requirements, even when it is applied to large-scale datasets, such as ImageNet; therefore, it can be employed in real-time applications that may also involve large-scale image spaces. The latter is also demonstrated in a drive-by experiment with clean and physically attacked traffic signs. Moreover, to evaluate the detection performance of VG against physical attacks, we  propose the first dataset, called AdvNet, with clean and physically attacked traffic sign images using the $\text{RP}_2$ attack \cite{eykholt2018robust}. We provide extensive comparisons that show that VG outperforms similar detectors \cite{meng2017magnet,feinman2017detecting,jha2018detecting} both in terms of scalability and detection performance. 

Finally, we would like to highlight that several white-box attacks have been proposed to bypass existing defenses that assume that the structure of the defense mechanism is fully known to the attacker \cite{papernot2016transferability,gilmer2018adversarial,carlini2017adversarial,carlini2017magnet}. Designing defenses against white-box attacks, although an important problem in this field, is out of the scope of this paper as their high computational complexity prohibits them from being applied to real-time scenarios that are of particular interest in this work. We would like to highlight again that our goal is to address an equally important issue which is to develop a computationally light defense mechanism that scales to large image domains, for real-time applications, a task that is particularly challenging for existing defenses both at design- and run-time as shown in our experiments. 
 
\subsection{Contribution}
The contributions of this paper can be summarized as follows. 
\textit{First}, we introduce VG, a new \textit{attack-agnostic} and \textit{dataset-agnostic} detection technique for defense against adversarial examples. \textit{Second}, we show that VG is more computationally efficient, both at run-time and design-time, than defenses that rely on training sets or building DNN-based detectors. This allows us to apply VG to real-time applications that may also involve large-scale image domains, illustrated by experiments on ImageNet.  \textit{Third}, we propose AdvNet, the first dataset with clean and physically attacked traffic sign images using the $\text{RP}_2$ attack \cite{eykholt2018robust} and the first evaluation against such robust physical attacks. \textit{Fourth}, we provide extensive comparative experiments on MNIST, CIFAR10, ImageNet, and AdvNet that show that VG outperforms similar defenses in terms of scalability and detection performance. 
%
%

\section{Prolem Statement: Detecting Adversarial Attacks}\label{sec:attacks}

%
Consider a classifier $f:\ccalX\rightarrow \ccalC$, where $\ccalX$ is the set of images $x\in\mathbb{R}^{n}$, where $n$ is the number of pixels, and $\ccalC$ is the set of labels. Let $L(x)$ and $f(x)$ denote the true and the predicted label of image $x\in\ccalX$, respectively. 
Then, the goal of an attacker is to perturb an image $x\in\ccalX$ by $\delta$ so that the difference between the perturbed and the original image is imperceptible and the perturbed image $x^*=x+\delta$ is missclassified, i.e., $f(x^*)\neq L(x)$.
%
%
In what follows, we provide a summary of existing digital and and physical adversarial attacks that can generate such perturbations $\delta$.

\subsection{Digital Attacks}\label{sec:dig}
{\bf{Fast Gradient Sign Method (FGSM):}} The Fast Gradient Sign Method (FGSM) \cite{goodfellow2014explaining} creates adversarial examples $x^*$ by perturbing the images $x$ in the direction the gradient of the loss function by magnitude $\epsilon$, where $\epsilon>0$ determines the perturbation size, i.e., 
\begin{equation}\label{eq:fgsm}
x^*=x+\epsilon\texttt{sign}(\nabla_x J(\theta, x,y)),
\end{equation}
where $\texttt{sign}(\cdot)$ is the sign function and $J(\theta, x,y)$ is the model's loss function with parameters $\theta$ and labels $y$. 


{\bf{Projected Gradient Descent (PGD):}} The Projected Gradient  Descent (PGD) method is a straightforward extension of FGSM. Specifically, it applies adversarial noise many times iteratively, giving rise to the following recursive formula: 
\begin{align}\label{eq:pgd}
&x_0^*=x,\nonumber\\
&x_i^*=\texttt{clip}_{\alpha,x}[x_{i-1}^*+\epsilon\texttt{sign}(\nabla_x J(\theta, \bbx_{i-1}^*))],
\end{align}
where $x_0^*=x$ and $\texttt{clip}_{\alpha,x}(\cdot)$ represent a clipping of the values of a sample so that it is within the $\alpha$-neighborhood of $x$. Compared to FGSM, this approach allows for extra control over the attack.

{\bf{Jacobian Saliency Map Attack (JSMA):}} An iterative method for targeted misclassification is proposed in \cite{papernot2016limitations}. Specifically, an adversarial saliency map is constructed based on the forward
derivative, as this gives the adversary the information required to make the neural network misclassify a given sample. For an input $x$ and a neural network $f$, the DNN output associated with the class $j$ is denoted by $f_j(x)$. To achieve a target class $t$, $f_t(x)$ must increase while the probabilities $f_j(x)$, where $j\neq t$ must decrease, until $t = \argmax_{c\in\ccalC} f_c(x)$.
The adversary can accomplish this by increasing input features using the
following saliency map $S(x, t)$
\begin{equation}
  \begin{split}
    S(x,t)[i] = \left\{ \begin{array}{ll} 0 &\text{if } g_t(x)<0 \\
      0 &\text{if } \sum_{j\neq t}g_j(x)>0 \\
      g_t(x)|\sum_{j\neq t}g_j(x)| &\text{otherwise},\nonumber\end{array} \right.
  \end{split}
\end{equation}
where $g_j(x) = \partial f_j(x)/\partial x_i$ and $i$ is  an input feature.  High values of $S(x, t)[i]$ correspond to input features
that will either increase the target class, or decrease other classes significantly, or both. Thus, the goal is to find input features $i, j$  that maximize $S(x,t)[i] + S(x,t)[j]$ and perturb these features by $\epsilon$. This process is repeated iteratively until the target misclassification is achieved.

{\bf{Carlini-Wagner (CW):}} An iterative (targeted or untargeted) attack to generate adversarial examples with small perturbations is proposed in \cite{carlini2017towards}. The perturbation $\delta$ is selected by solving the following optimization problem:
\begin{align}\label{eq:cw}
&\underset{\delta}{\text{minimize}}
 \;\;\; ||\delta||_p+\epsilon g(x+\delta)\\
&\text{subject to} ~~~x+\delta\in[0,1]^n,\nonumber
\end{align}
where $\epsilon\geq 0$ is a suitably chosen constant and $g(x+\delta)$ depends on the softmax output of the neural network and is selected so that $g(x+\delta)\leq 0$ if the perturbed image is misclassified or gets a desired label, and $g(x+\delta)> 0$ otherwise. 

\subsection{Robust Physical Perturbations ($RP_2$)}\label{sec:RP2}
 
A targeted robust physical attack is presented in \cite{eykholt2018robust} to generate robust visual adversarial perturbations under different physical conditions. The first step requires to solve the following optimization problem that generates adversarial (digital) noise $\delta$:
\begin{align}\label{eq:physical}
&\underset{\delta}{\text{minimize}}
 \;\;\; \lambda||M_x \delta||_p+NPS+\\
&~~~~~~~~~\mathbb{E}_{x_i\sim X^V}J(f(x_i+T_i(M_x\delta)),y^*),\nonumber
\end{align}
where (i) $M_x$ is a mask applied to image $x$ to ensure that the perturbation is applied only to the surface of the object of interest (e.g., on a traffic sign and not in the background); (ii) NPS is a non-printability score to account for fabrication error; (iii) $X^V$ refers to a distribution of images containing an object of interest (e.g., a stop sign) under various environmental conditions captured by digital and physical transformations (resulting in attacks being robust to various environmental conditions) ; (iv)  $T_i(\cdot)$ denotes the alignment function that maps transformations on the object to transformations on
the perturbation (e.g. if the object is rotated, the perturbation is rotated as well); and (v) $y^*$ is the target label. 

\begin{figure}[t]
  \centering
    \includegraphics[width=1\linewidth]{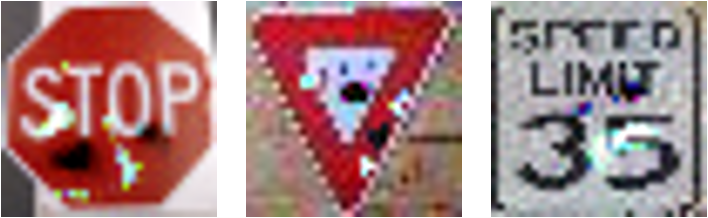}
  \caption{Perturbation generated by the $\text{RP}_2$ attack using the authors-provided code under the $\ell_1$ norm so that a stop, yield, and speed limit 35 sign are misclassified by the LISA CNN \cite{eykholt2018robust} as speed limit 35, speed limit 35, and turn right sign, respectively.}
  \label{fig:attackedl1}
\end{figure}

Finally, an attacker will print out the optimization result on paper, cut out the perturbation $M_x$, and put it onto the target object. The perturbation generated using the $L_1$ norm along with its physical application is shown in Figure \ref{fig:attackedl1}.  Observe in this figure, that the $L_1$ norm generates a sparse attack vector allowing the attacker to physically implement the attack with black and white stickers. For instance, application of the perturbation for the stop sign in the form of black and white stickers was shown in Figure \ref{fig:attack_example_p}. Generation of physical stickers for the yield and speed limit sign is shown in Section \ref{sec:sim} (see Figure \ref{fig:X}).

The goal in this paper can be summarized as follows:

\begin{problem}
Given a DNN $f:\ccalX\rightarrow\ccalC$ that is subject to the digital and physical attacks discussed in Sections \ref{sec:dig}-\ref{sec:RP2}, design a detector $f_d:\ccalX\rightarrow\{\text{adversarial, clean}\}$ that classifies input images to $f$ as adversarial or clean.
\end{problem}

The main assumption that we make throughout this paper is that both the the attacker and the defender have full knowledge of the DNN $f$. However, the attacker is oblivious to the detector/defense mechanism and the defender is not aware of which specific attack (e.g., CW or FGSM) the DNN is subject to. 



\section{VisionGuard: A New Image Defense Framework}\label{sec:detector}
\begin{figure}[t]
  \centering
    \includegraphics[width=1\linewidth]{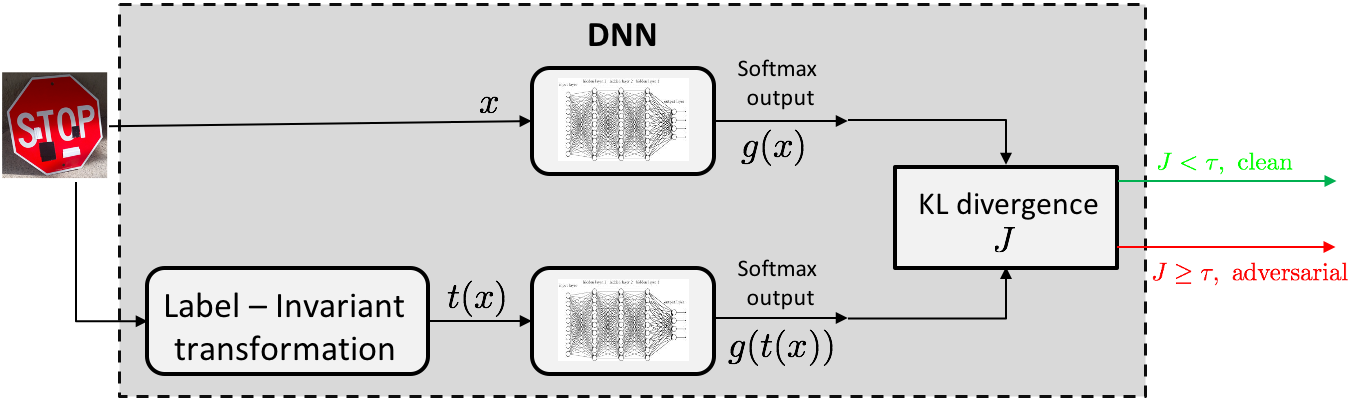}
  \caption{Graphical illustration of VisionGuard (gray box). VisionGuard classifies an input image under investigation as clean or adversarial using label-invariant transformations and the structure of the DNN under attack.}
  \label{fig:vg}
\end{figure}

Our goal is to build a detector $f_d:\ccalX\rightarrow\{0,1\}$, such that (i) $f_d(x)=0$ if the image $x$ is a legitimate image and (ii) $f_d(x)=1$ if $x\in\ccalX$ is an adversarial input, i.e., if it has been manipulated/perturbed. In what follows, we propose VisionGuard (VG), an attack-agnostic transformation-based detector; see also Algorithm \ref{alg:detect} and Figure \ref{fig:vg}.  

\begin{algorithm}[t]
   \caption{VisionGuard}
   \label{alg:detect}
\begin{algorithmic}
   \STATE {\bfseries Input:} \{Input image $x$, DNN $f:\mathbb{R}^n\rightarrow\ccalC$, transformation $t$\}
   \STATE {\bfseries Output:} \{$f_d(x)=1$ if $x$ is an adversarial input, and $f_d(x)=0$ otherwise\}
   \STATE Feed $x$ to the DNN model and get softmax output $\bf{g}(x)$\label{det:gx}
   \STATE Apply transformation $t$ to $x$ and get image $x'=t(x)$.
   \STATE Feed $x'$ to the DNN model and get softmax output $\bf{g}(x')$
   \STATE Compute $J(x,x')=\min(D_{\text{KL}}(\bbg(x),\bbg(x')),D_{\text{KL}}(\bbg(x'),\bbg(x)))$\;\label{det:KL}
   \STATE Initialize $f_d(x)=0$
   \IF{$J(x,x')\geq \tau$}
   \STATE $f_d(x)=1$
   \ENDIF
\end{algorithmic}
\end{algorithm}
%
%
The proposed detector relies on the observation that adversarial inputs are not robust to certain transformations in the sense that they change the output of the DNN significantly. Typically, such transformations are (i) label-invariant, i.e., the true class of the object of interest should not change under this transformation, and (ii) squeeze out features that may be unnecessary for correct classification such as adversarial components. Examples of such transformations are lossy-compression, zoom-in, zoom-out, brightness, and cropping; similar transformations have been used in image purification domain to `remove' digital adversarial noise from images but in a completely different way. For instance, lossy compression (e.g., JPEG compression) may remove digital adversarial noise while zooming out and brightness transformations may alleviate the effect of physical attacks; see Section \ref{sec:sim}.

VG comprises four steps to detect whether an input image $x$ is adversarial or not. First, the input image $x$ is fed to the classifier $f$ to get the softmax output denoted by $\bbg(x)$. Second, a user-specified transformation $t:\ccalX\rightarrow\ccalX$ is applied to $x$ to get an image $x'=t(x)$. Third, $x'$ is fed to the classifier to get the softmax output denoted by $\bbg(x')$. Fourth, $x$ is classified as adversarial if the softmax outputs $\bbg(x)$ and $\bbg(x')$ are significantly different. Formally, we measure similarity between $\bbg(x)$ and $\bbg(x')$ using the K-L divergence measure, denoted by $D_{\text{KL}}(\bbg(x),\bbg(x')$ and defined as follows:
\begin{equation}
    D_{\text{KL}}(\bbg(x),\bbg(x') = \sum_{c\in\ccalC}\bbg_c(x)\log\frac{\bbg_c(x)}{\bbg_c(x')}
\end{equation}
where $\bbg_c(x)$ denotes the $c$-th entry in the softmax output vector $\bbg(x)$; in other words, $\bbg_c(x)$ can be viewed as the probability that the class of image $x$ is $c$.  Specifically, if $$J(x,x')=\min(D_{\text{KL}}(\bbg(x),\bbg(x')),D_{\text{KL}}(\bbg(x'),\bbg(x)))$$ is greater than a threshold $\tau$, then $x$ is considered an adversarial input, i.e., $f_d(x)=1$; otherwise, $x$ is classified as a legitimate image, i.e., $f_d(x)=0$. 

%



{\bf{Detection Thresholds:}} To determine the detection threshold $\tau$ we use Receiver Operating Characteristics (ROC) graphs that are constructed as follows. 
First, given a set $\ccalX$ of clean images we construct the corresponding set of adversarial images, denoted by $\ccalX_a$, using any attack or possibly a mixture of attacks. Next, recall that our detection mechanism $f_d$ maps each image to $0$ (legitimate input) or $1$ (adversarial input). Hereafter, we call the class of adversarial images as `positives' and the class of clean images as `negatives'. Then, given a threshold $\tau$, we estimate the true positive rate as the number of true positives (i.e., the number of adversarial inputs classified as adversarial inputs) divided by the total number of positives, i.e., the total number of adversarial images. Similarly, we estimate the false positive rate as the number of false positives (i.e., the number of legitimate inputs classified as adversarial inputs) divided by the total number of negatives, i.e., the total number of clean images. Then, ROC graphs can be constructed by plotting the TP rate on the Y axis and the FP rate on the X axis for various thresholds $\tau$. Given an ROC graph, we select the threshold that returns the closest point to $(0, 1)$, since this point corresponds to perfect attack detection.

\section{Experiments}\label{sec:sim}

\begin{figure}[t]
  \centering
     \subfigure[FGSM, PGD, and JSMA]{
    \label{fig:acc1}
    \includegraphics[width=0.8\linewidth]{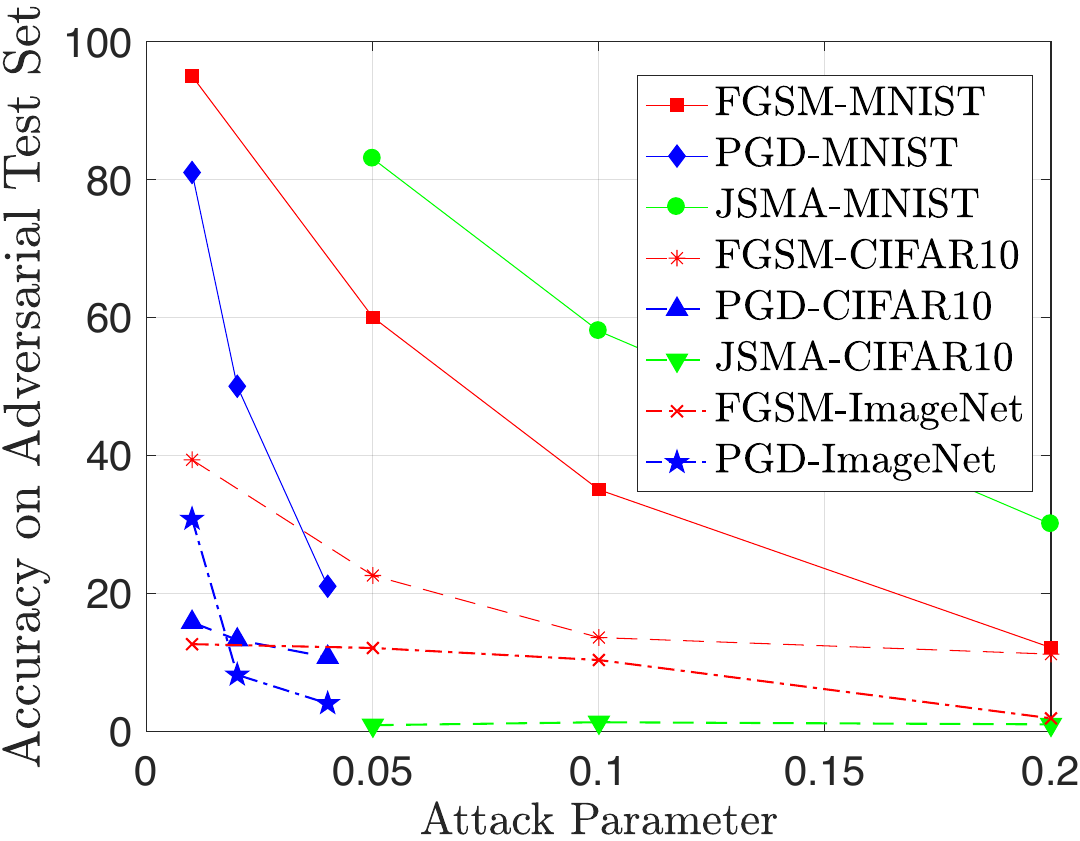}}
  \subfigure[CW]{
    \label{fig:acc2}
    \includegraphics[width=0.8\linewidth]{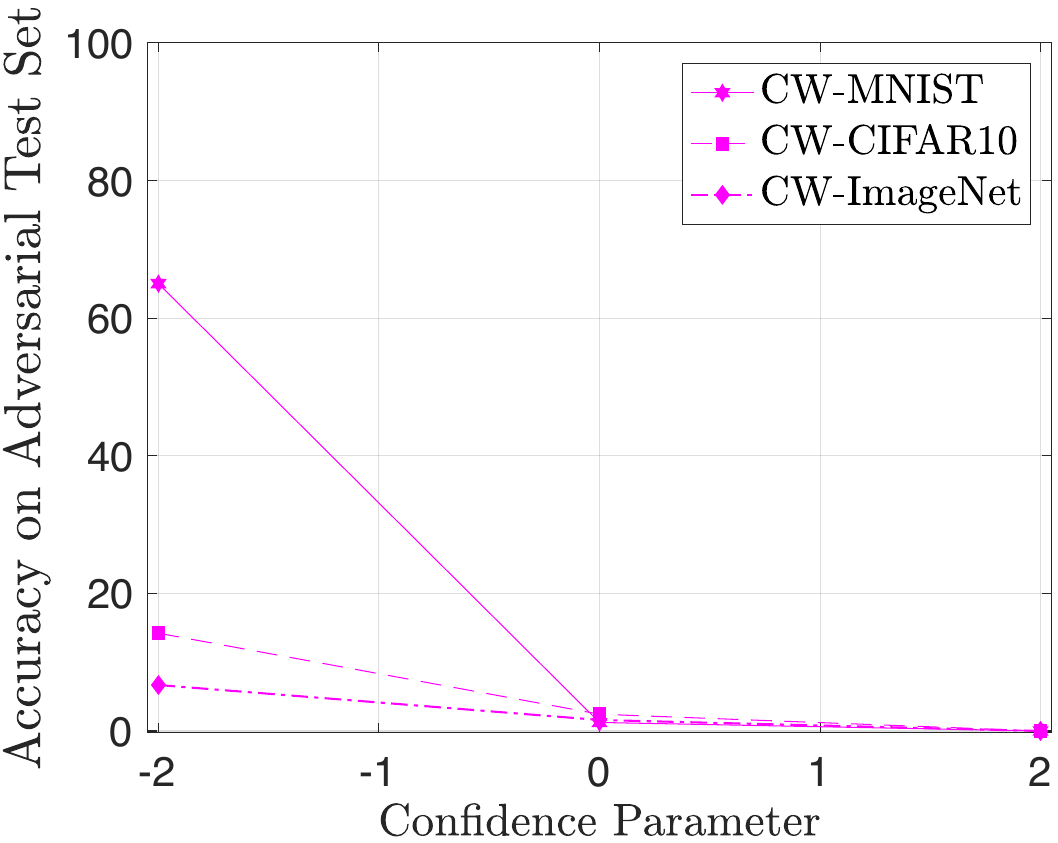}} 
  \caption{
  Accuracy of the target DNN classifier on test sets of adversarial images generated by FGSM, PGD, JSMA, and CW for various attack-specific parameters $\epsilon$ on the MNIST, CIFAR-10, and ImageNet datasets. 
  }
  \label{fig:accuracy}
\end{figure}

\begin{table}[t]
\caption{Runtime (secs) of Attack Algorithms per Image}
\centering
\label{tab:runtimeAtt}
\begin{tabular}{|l|l|l|l|}
\hline
     & MNIST   & CIFAR10 & ImageNet \\ \hline
FGSM & 0.00005 & 0.0017  & 0.1728        \\ \hline
PGD  & 0.005   & 1.11    & 8.46        \\ \hline
JSMA & 1.75    & 19.67   & DNF     \\ \hline
CW   & 0.005   & 1.01    & 16.39    \\ \hline
\end{tabular}
\end{table}

\begin{table}[t]
\centering
\caption{MNIST: Comparative experiments in terms of AUC against MagNet and KDE.}
\label{tab:MNIST}
\begin{tabular}{|l|l|l|l|}
\hline
MNIST             & VG+JPEG92 & MagNet & KDE   \\ \hline
FGSM $\epsilon=0.01$ &  \textbf{61.6\%}     &  58.4\%  & 56.2\% \\ \hline
FGSM $\epsilon=0.05$ & 75.9\%     &  71.8\%  & \textbf{76.5\%} \\ \hline
FGSM $\epsilon=0.1$  &  71.3\%     &  72.9\%  & \textbf{84.5\%} \\ \hline
FGSM $\epsilon=0.2$ & 57.1\%     & 51.6\%  & \textbf{92.0\%} \\ \hline
PGD $\epsilon=0.01$ &  \textbf{93.7\%}     &  87.5\%  & 83.7\% \\ \hline
PGD $\epsilon=0.02$  &  \textbf{90.0\%}     &  85.9\%  & 81.7\% \\ \hline
PGD $\epsilon=0.04$  & 79.8\%     & 76.7\%  & \textbf{85.5\%} \\ \hline
JSMA $\epsilon=0.05$ &  \textbf{71.7\%}     & 71.3\%  & 65.7\% \\ \hline
JSMA $\epsilon=0.1$  & \textbf{83.4\%}     &  80.2\%  & 77.8\% \\ \hline
JSMA $\epsilon=0.2$  & \textbf{91.7\%}     &  85.5\%  & 85.8\% \\ \hline
CW $\epsilon=-2$     & \textbf{96.3\%}     & 95.7\%  & 88.1\% \\ \hline
CW $\epsilon=0.01$  &  \textbf{96.1\%}     &  96.0\%  & 89.9\% \\ \hline
CW $\epsilon=2$     &  94.7\%     &  \textbf{96.0\%}  & 87.3\% \\ \hline
\end{tabular}
\end{table}
\normalsize

\begin{table}[t]
\centering
\caption{CIFAR10: Comparative experiments in terms of AUC against MagNet and KDE}.
\label{tab:cifar10}
\begin{tabular}{|l|l|l|l|}
\hline
CIFAR10           & VG+JPEG92 & MagNet & KDE   \\ \hline
FGSM $\epsilon=0.01$ & 84.1\%     &  \textbf{84.3\%}  & 80.8\% \\ \hline
FGSM $\epsilon=0.05$ & 82.9\%     & \textbf{89.8\%}  & 89.2\% \\ \hline
FGSM $\epsilon=0.1$  & 78.3\%     & 89.7\%  & \textbf{90.4\%} \\ \hline
FGSM $\epsilon=0.2$  & 76.9\%     & 85.5\%  & \textbf{89.7\%} \\ \hline
PGD $\epsilon=0.01$  & 93.7\%     &  93.9\%  & \textbf{96.9\%} \\ \hline
PGD $\epsilon=0.02$  &  91.1\%     &  93.7\%  & \textbf{96.4\%} \\ \hline
PGD $\epsilon=0.04$  &  89.2\%     & 93.4\%  & \textbf{95.9\%} \\ \hline
JSMA $\epsilon=0.05$ &  93.9\%     & 93.4\%  & \textbf{95.0\%} \\ \hline
JSMA $\epsilon=0.1$  & 94.2\%     & 93.7\%  & \textbf{95.5\%} \\ \hline
JSMA $\epsilon=0.2$  &  94.2\%     &  93.7\%  & \textbf{95.6\%} \\ \hline
CW $\epsilon=-2$     &  87.2\%     & \textbf{90.0\%}  & 89.6\% \\ \hline
CW $\epsilon=0.01$  & 85.8\%     & \textbf{89.5\%}  & 89.0\% \\ \hline
CW $\epsilon=2$      & 84.8\%     & \textbf{88.9\%}  & 87.1\% \\ \hline
\end{tabular}
\end{table}


\begin{table*}[t]
\centering
\caption{ImageNet: Comparative experiments in terms of AUC against MagNet and KDE}.
\label{tab:imagenet}
\begin{tabular}{|l|l|l|l|l|l|l|}
\hline
ImageNet          & VG+JPEG75 & VG+JPEG92 & VG+JPEG98 & VG+Median3 & MagNet                                                                                  & KDE    \\ \hline
FGSM $\epsilon=0.01$ & 84.0\%     & \textbf{89.7\%}     & 86.6\%     & 85.0\%      & \multirow{10}{*}{\begin{tabular}[c]{@{}l@{}}~~~~~~~~DNF\\ (could not train \\ auto-encoders)\end{tabular}} & 47.1\% \\ \cline{1-5} \cline{7-7} 
FGSM $\epsilon=0.05$ & 85.3\%     & \textbf{94.7\%}     & 86.7\%     & 88.9\%      &                                                                                         & 47.9\% \\ \cline{1-5} \cline{7-7} 
FGSM $\epsilon=0.1$  & 88.9\%     & \textbf{98.7\%}     & 88.8\%     & 93.3\%      &                                                                                         & 46.0\% \\ \cline{1-5} \cline{7-7} 
FGSM $\epsilon=0.2$  & 91.0\%     & \textbf{99.8\%}     & 82.2\%     & 94.0\%      &                                                                                         & 47.4\% \\ \cline{1-5} \cline{7-7} 
PGD $\epsilon=0.01$  & 89.6\%    & \textbf{94.5\%}     & 87.4\%     & 86.7\%       &                                                                                         & 43.2\% \\ \cline{1-5} \cline{7-7} 
PGD $\epsilon=0.02$  & 89.8\%     & \textbf{98.7\%}     & 93.4\%     & 87.8\%      &                                                                                         & 46.8\% \\ \cline{1-5} \cline{7-7} 
PGD $\epsilon=0.04$  & 94.3\%     & \textbf{98.4\%}     & 91.2\%     & 80.1\%       &                                                                                         & 47.6\% \\ \cline{1-5} \cline{7-7} 
CW $\epsilon=-2$    & 93.2\%    & \textbf{97.9\%}     & 82.6\%     & 82.4\%      &                                                                                         & 54.9\% \\ \cline{1-5} \cline{7-7} 
CW $\epsilon=0$      & 87.2\%     & \textbf{91.9\%}     & 88.3\%     & 88.6\%      &                                                                                         & 47.4\% \\ \cline{1-5} \cline{7-7} 
CW $\epsilon=2$     & 80.9\%     & 85.6\%     & 85.0\%     & \textbf{87.9\%}      &                                                                                         & 47.2\% \\ \hline
\end{tabular}
\end{table*}

First, in Section \ref{sec:digital}, we tested VG against the state-of-the-art digital attacks, FGSM, PGD, JSMA, and CW, on three standard machine learning datasets: MNIST, CIFAR 10, and ImageNet. Comparisons against the KDE \cite{feinman2017detecting,jha2018detecting} and MagNet \cite{meng2017magnet} detectors are also presented showing that VG outperforms both in terms of scalability and detection performance.
Second, in Section \ref{sec:physical} we evaluated VG against the robust physical attack $\text{RP}_2$ on AdvNet, a dataset that we constructed consisting of clean and physically attacked traffic signs under various environmental conditions. Finally, to illustrate the need for a real-time detector, we also evaluated VisionGuard on data collected in a drive-by experiment conducted in \cite{eykholt2018robust}. VisionGuard achieves high detection performance on both AdvNet and the drive-by experiment while outperforming the KDE detector. Finally, in Section \ref{sec:realtime}, we evaluated the real-time computational requirements of VG, KDE, and MagNet in terms of required disk-space and execution time. All experiments have been executed on a computer with Intel(R) Xeon(R) Gold 6148 CPU, 2.40GHz.



\subsection{Evaluation Against Digital Attacks}\label{sec:digital}
In this section, first we evaluate how effective digital attacks are, i.e., how much they can drop the accuracy of DNNs on the MNIST, CIFAR10, and ImageNet datasets. Second, we evaluate the detection performance of VG, KDE, and MagNet against digital attacks. A discussion about generalization of these detectors across datasets is also provided.

{\bf{Description of Datasets and Target DNNs:}} MNIST contains $70,000$ grayscale $28\times 28$ images  divided into $60,000$ training samples and $10,000$ test samples with $10$ classes. CIFAR10 contains $60,000$ RGB $32\times 32$ images divided into $50,000$ training samples and $10,000$ test samples with $10$ classes, as well. ImageNet (ILSVRC2012) contains $1.4$ million RGB $224\times 224$ images divided in $1.2$ million training images, $50,000$ validation images, $150,000$ testing images, with $1000$ classes. For the purposes of experimentation, we treat the validation set as the training set due to the availability of labels.
For the MNIST classification task, we consider a simple, fully-connected neural network with two hidden layers and $64$ neurons per layer that achieves $97.2\%$ accuracy on the test set. As for the CIFAR10 and ImageNet datasets, we consider convolutional neural networks with residual blocks (ResNet-56 and ResNet-50, respectively) \cite{he2016deep}. The accuracy of the trained ResNet-56 and ResNet-50 is $93.7\%$ and $73.4\%$, respectively.

{\bf{Evaluation of Digital Attacks:}} We apply the FGSM, PGD, JSMA, and CW attacks, for various attack-specific parameters $\epsilon$ (see Section \ref{sec:dig}) on the MNIST, CIFAR10, and ImageNet test sets. Note that for the PGD attack, we select $a=10\epsilon$; see \eqref{eq:pgd}. The accuracy of the classifier on the resulting adversarial test sets is depicted in Figure \ref{fig:accuracy}. 
Observe in this figure that as the magnitude of the perturbation increases, the accuracy of the neural network decreases.  Moreover, observe that as the image dimensions increase, it is easier to fool the target classifier. The reason is that adversaries can search for adversarial inputs over larger input feature spaces.
In Table \ref{tab:runtimeAtt}, we also report the average runtime required to generate a single adversarial image on MNIST, CIFAR10, and ImageNet using the FGSM, PGD, JSMA, and CW attacks and code available online by the authors. Observe that the most computationally-light attack is FGSM while the most computationally-expensive is JSMA. In fact, JSMA failed to generate an adversarial ImageNet image due to memory constraints within the attack method. Also, note that there is a trade-off between computational-efficiency and effectiveness of the attack. Specifically, observe in Figure \ref{fig:accuracy} that e.g., on CIFAR10, FGSM and JSMA are the most and least effective attacks, respectively.



{\bf{Evaluation of VisionGuard \& Comparative Experiments:}} In what follows, we evaluate the efficacy of VG using ROC graphs. 
For the construction of ROC graphs, we call `positives' the images (i) that have been attacked (even if the attack fails, i.e., it does not cause misclassification) and (ii) the clean images that are misclassified as they can also been seen as `adversarial' inputs. All other images (i.e., clean images that are correctly classified) are called 'negatives'. We examine the performance of VG when it is integrated with JPEG compression with various compression qualities and with median filters. 
Note that VG along with rotation transformations, such as the ones used in \cite{tian2018detecting}, or bit-depth reduction transformations, as used in \cite{guo2017countering,das2018shield}, yield poor detection performance and, therefore, such results are omitted. Finally, we provide comparisons against MagNet \cite{meng2017magnet} and the KDE detector that is originally proposed in \cite{feinman2017detecting,jha2018detecting}. 
To compare against MagNet and KDE, we leverage the code provided by the authors.

{\textit{MNIST:}} 
Table \ref{tab:MNIST} presents the area under the ROC graphs (AUC) when VG is applied using JPEG compression with $92\%$ compression quality. Similar performance was seen for compression qualities $75\%$, $92\%$, $98\%$, and median $3 \times 3$ filter. The additional results are omitted due to space limitations.  
Observe in Table \ref{tab:MNIST} that VG outperforms both MagNet and KDE in almost all attacks. Also, note that VG and MagNet fail to detect FGSM-generated adversarial inputs.

{\textit{CIFAR10:}} The respective AUC comparison for CIFAR10 is presented in Table \ref{tab:cifar10}. VG, MagNet, and KDE have comparable AUC-based performance on adversarial images generated using the PGD, JSMA and CW attacks, while both KDE and MagNet outperform VG on FGSM-based adversarial inputs, especially for large values of the attack parameter $\epsilon$. Note that \cite{carlini2017adversarial} states that the KDE detector gives poor performance on CIFAR10, which contradicts our results. 

{\textit{ImageNet:}} To evaluate VG on ImageNet, we have randomly sampled $2,000$ images; the results are summarized in Table \ref{tab:imagenet}. Observe that VG attains high AUC-based performance ($>90\%$) for almost all attacks and any attack parameters. MagNet requires a new auto-encoder for each new dataset it is applied to. An auto-encoder for ImageNet is not provided by the authors in \cite{meng2017magnet} while training such an auto-encoder did not finish within two weeks; therefore, comparisons on ImageNet are not available. Furthermore, extracting the embeddings for $1.2$ million ImageNet images, as required in \cite{feinman2017detecting,jha2018detecting,pang2018towards}, required $36$ hours approximately. 
%
In Table \ref{tab:imagenet} we report the performance of the KDE detector which is comparable to the performance of a random detector.

%

{\bf{Dataset-agnosticity:}} Recall that MagNet requires a new autoencoder for each new dataset it is applied to. Similarly, the KDE-based detectors \cite{feinman2017detecting,jha2018detecting,pang2018towards} rely on extracting embeddings from training sets. As a result, it is questionable if these detectors achieve high detection performance when they are deployed in real-world environments for which datasets may not exist. Notice in Tables \ref{tab:MNIST}-\ref{tab:imagenet} that VisionGuard is dataset-agnostic as the same transformation, JPEG with compression quality $92\%$ yields high detection performance across the MNIST, CIFAR10, and ImageNet datasets that differ both in content and image dimensions.

%
 

{\bf{Robustness to Random Noise:}} As JPEG compression is sensitive to noise, we also evaluated the robustness of VG on random noise. Specifically, first we add random Gaussian noise to the original clean images. The accuracy of the classifiers on the generated noisy MNIST, CIFAR10, and ImageNet datasets dropped by $1\%$ on average across all datasets indicating that the generated noise caused misclassifications. 
Then, we compute the ROC curves where positives are the original clean and correctly classified images and negatives are the resulting noisy and correctly classified images. The AUC is $53.65\%$,  $55.35\%$, and  $44.62\%$  on MNIST, CIFAR10, and ImageNet, respectively. This shows that VG cannot distinguish between clean and noisy images and, therefore, it will not raise false alarms due to random noise, e.g., dust in the camera lens.

\subsection{Evaluation Against Physical Attacks}\label{sec:physical}
In this section, first we describe AdvNet, our proposed dataset consisting of clean and physically attacked traffic signs using the robust physical attack $\text{RP}_2$ \cite{eykholt2018robust}. To generate the physical stickers, we leverage the code provided by the authors. Second, we evaluated the proposed detector on AdvNet and on data collected in a drive-by experiment conducted in \cite{eykholt2018robust} showing that VG achieves high detection performance while outperforming the KDE detector.

\begin{figure}[t]
  \centering
  \subfigure[]{
    \label{fig:stop2}
    \includegraphics[width=0.30\linewidth]{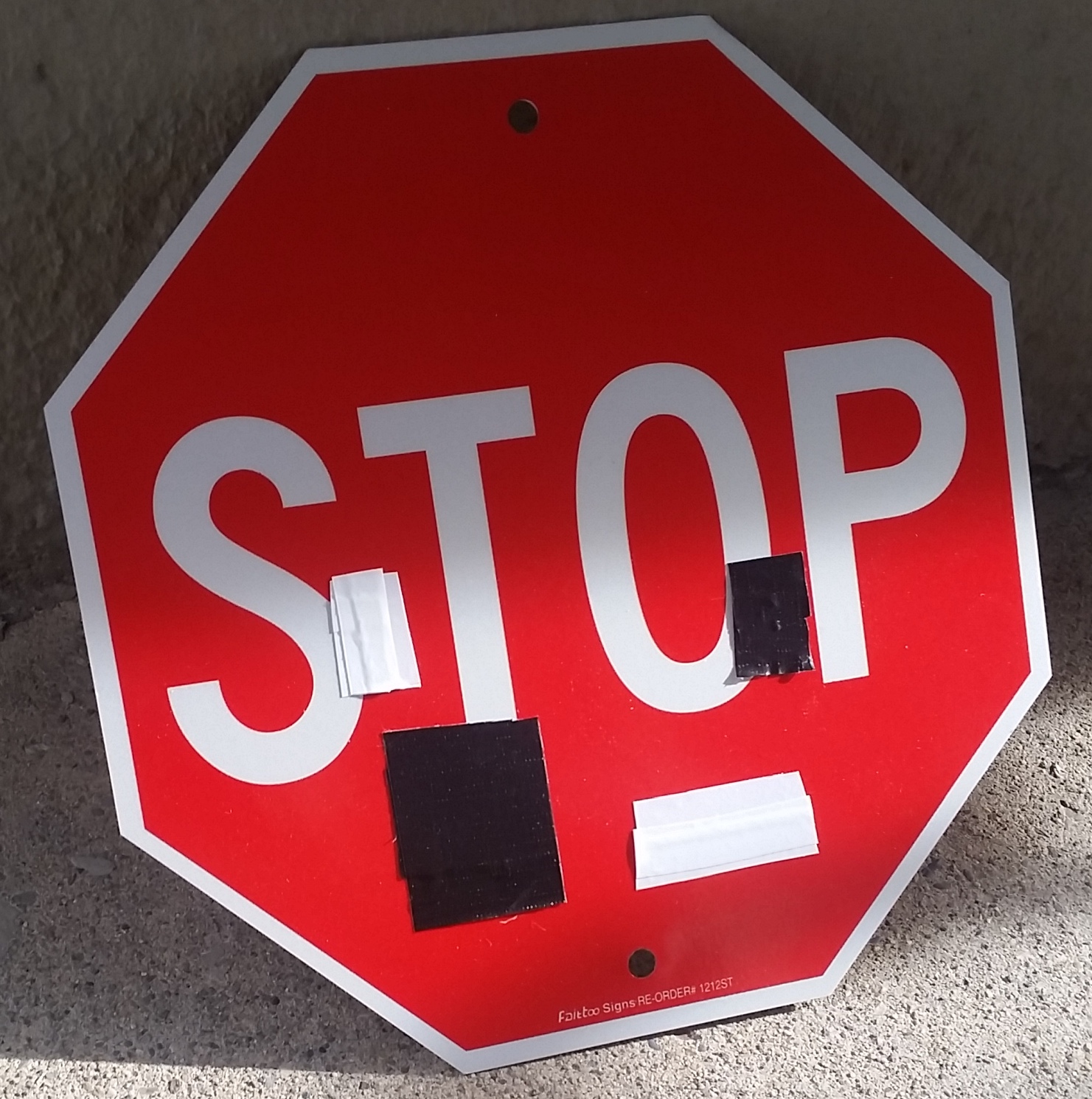}}
     \subfigure[]{
    \label{fig:yield1}
    \includegraphics[width=0.30\linewidth]{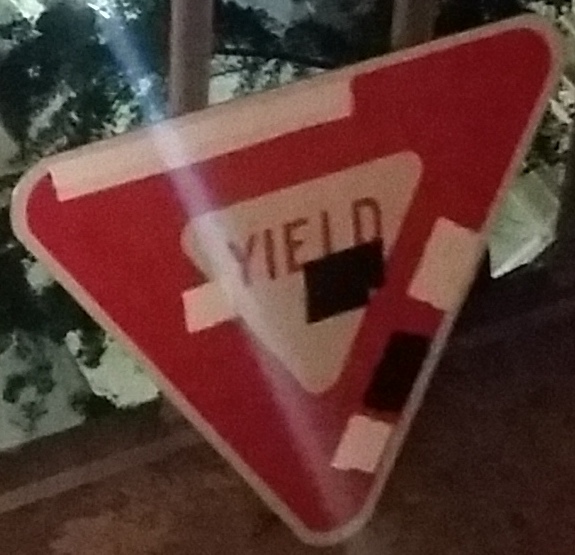}}
       \subfigure[]{
    \label{fig:speed2}
    \includegraphics[width=0.22\linewidth]{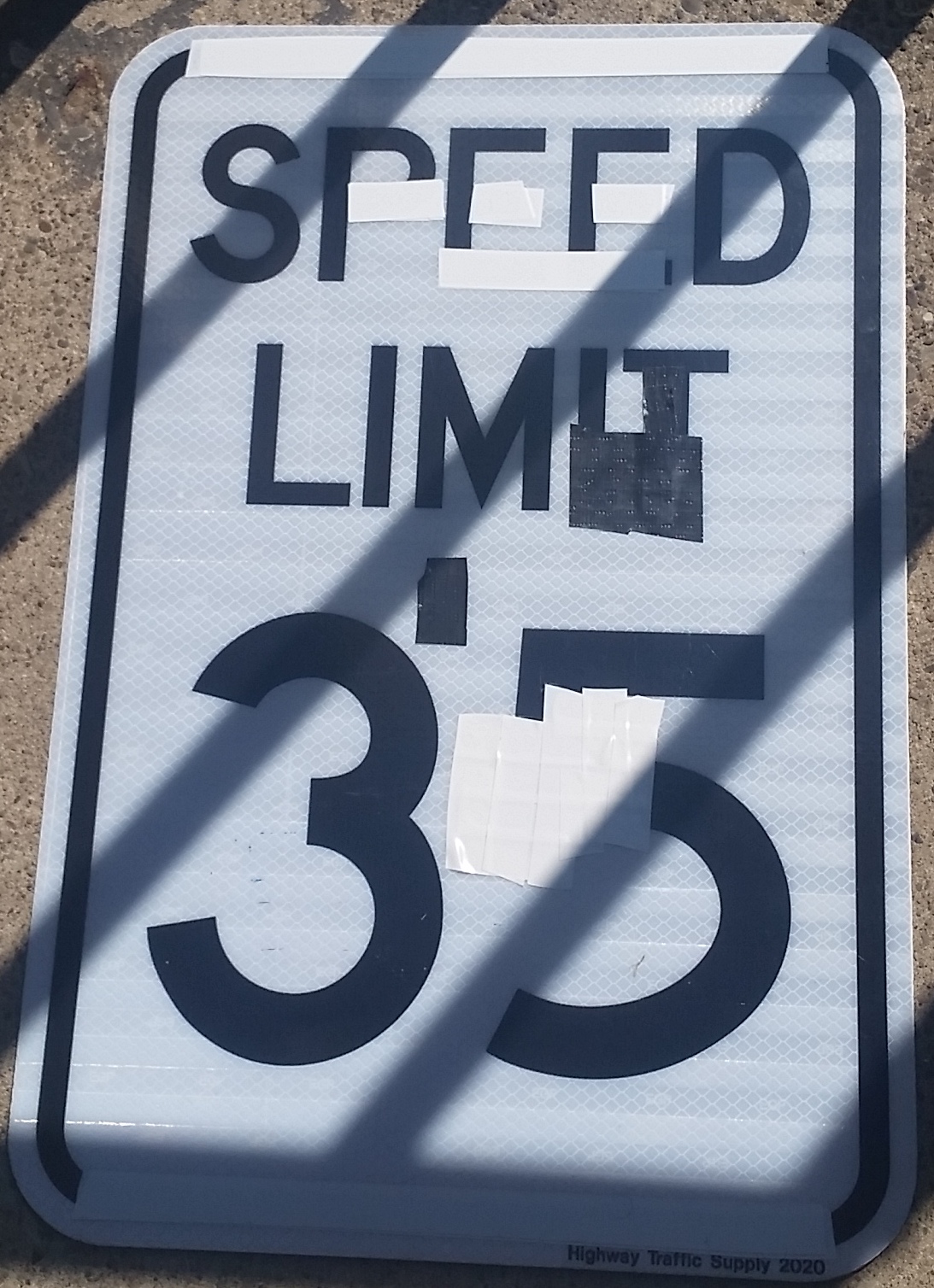}}
  \caption{
   Traffic signs manipulated by adversarial stickers generated as per the $\text{RP}_2$ attack \cite{eykholt2018robust} to fool LISA-CNN. The stickers have been placed as per the perturbation shown in Figure \ref{fig:attackedl1}.
  }
  \label{fig:X}
\end{figure}

{\bf{AdvNet, Target DNNs, and Evaluation of $\text{RP}_2$:}} 
%
%
To evaluate the proposed detector against the physical attack $\text{RP}_2$, we have collected both clean and adversarial images for the following traffic signs: 'stop', 'speed limit 35', and 'yield'. All images have been taken with a $12$MP smartphone camera under various environmental conditions (e.g., lighting, angle, distance, and background); see e.g., Figure \ref{fig:X}. The adversarial images are generated by the $\text{RP}_2$ attack using the $\ell_1$ norm. Specifically, given an image with an object of interest (e.g., a stop sign) and a target label (e.g., a speed limit 35 sign) we execute the $\text{RP}_2$ attack to generate the corresponding adversarial image that can fool the target neural network. The resulting (digital) adversarial image is used as a guide to place black and white stickers in the physical world, on the surface of the object of interest, to fool the target perception system. Recall that this attack depends on the target neural network since effective physical attacks differ across DNNs. Hereafter, we consider the LISA-CNN \cite{eykholt2018robust} with
91\% accuracy on the LISA test set and GTSRB-CNN \cite{stallkamp2012man} with 95.7\% accuracy on the GTSRB test set. Therefore, we have generated adversarial images for each CNN. Specifically, AdvNet contains images with clean stop signs, clean speed limit 35 signs, clean yield signs, adversarial stop signs for LISA-CNN, adversarial stop signs for GTSRB-CNN, adversarial yield signs for LISA-CNN, and adversarial speed limit signs for LISA-CNN. In total, AdvNet consists of $2,645$ clean and and $4,007$ adversarial traffic sign images collected under various environmental conditions that include angle, distance, background, and lightning conditions. 

In particular, for the LISA CNN, the stop, yield, and speed limit 35 sign are attacked so that they are missclassified as speed limit 35, speed limit 35, and turn right signs, respectively. The accuracy of the generated sticker attack (e.g., the percentage of adversarial stop signs classified as speed limit 35 signs) on stop, yield, and speed limit 35 sign is $3.7\%$, $48.18\%$, $48.55\%$. Note that for the stop signs we placed stickers as shown in \cite{eykholt2018robust}. Nevertheless, in our implementation, only $3.7\%$ of the adversarial stop signs were classified as speed limit 35 signs while $36.76\%$ of them were classified as as speed limit 55 signs. This may show that the adversarial stickers may also depend on other factors not considered in \cite{eykholt2018robust} such as the background. Also, we would like to highlight that in our experiments, we realized that small changes in the size and location of the stickers also affect the accuracy of the attack which may have contributed to the above result too.

As for the GTSRB-CNN, recall that it is trained on the German Traffic Sign Recognition Benchmark. Since we did not have access to German traffic signs for our physical experiments, GTSRB-CNN has been evaluated only on US stop signs. The stickers are placed as in \cite{eykholt2018robust} so that the GTSRB-CNN classifies stop signs as speed limit 35 signs. In our implementation, $0\%$ of the collected adversarial images is missclassified as a speed limit sign; instead $44.71\%$ of these images are classified as pedestrian crossing signs. 

In total, the constructed dataset consists of clean stop signs, clean yield signs, clean speed limit 35 signs, adversarial stop signs for GTSRB-CNN, adversarial stop signs for LISA-CNN, and adversarial yield and speed limit signs for LISA-CNN with approximately $1000$ images with dimensions $32\times32\times 3$ per label for the clean and adversarial set. The accuracy of both CNNs on the collected clean and adversarial images can be found in Table \ref{tab:accRP2} showing that the $\text{RP}_2$ attack results in decreasing the accuracy of the CNNs by at least $50\%$. Examples of the generated adversarial images are shown in 
Figure \ref{fig:X}.
%


\begin{figure}[t]
  \centering
    \label{fig:stopTransf}
    \includegraphics[width=1\linewidth]{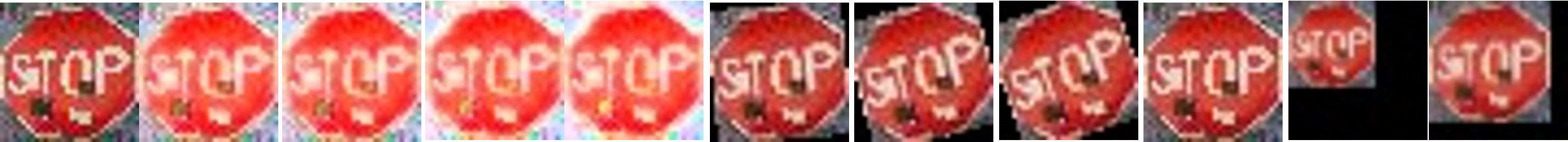}
\caption{Examples of transformed images. The leftmost picture corresponds to the original picture while the rest are transformed versions using B90, B110, B150, B200, R5, R15, R20, JPEG92, Z20, and Z28 (from left to right).}
  \label{fig:transf}
\end{figure}

{\bf{Evaluation of VisionGuard \& Comparative Experiments:}} In what follows, we evaluate the efficacy of VG on the $\text{RP}_2$ attack using ROC graphs. 
We examine the performance of VG when it is integrated with JPEG compression with compression quality $92\%$, brightness, rotation, and zoom-out transformations. The results are summarized for LISA-CNN and GTSRB-CNN in Tables \ref{tab:LISAall}-\ref{tab:gtsrb}. In these tables, $Bd$ and $Rd$ refer to brightness transformation with HSV value equal to $d$ and $d$-degree rotation, respectively. The larger the value of $d$ is in $Bd$, the brighter the image. Also, in the same table, $Zd$ refers to the zoom-out transformation performed as follows. First, the image is resized so that its dimensions are $d\times d \times 3$, where $d<32$. The resized image is augmented with black pixels so that a valid input to the CNNs is created with dimensions $32\times 32\times 3$. Examples of transformed images are shown in Figure \ref{fig:transf}. Observe in this figure that the brightness transformation removes features introduced by the attacker. Specifically, the brighter the image is, the less visible the black stickers are.

For both CNNs, the transformation that yields the best detection performance is brightness with HSV equal to $200$ yielding an AUC equal to $89.43\%$ outperforming the KDE detector; see Table \ref{tab:LISAall}. Specifically, the latter achieves AUC equal to $81.56\%$. In this comparison, the classifier required in the KDE detector is trained using a training that collects $80\%$ of all images per label. The reported AUC was computed based on the remaining AdvNet images. 
Observe also in Table~\ref{tab:LISAall} that the lossy compression transformation, employed for the digital attacks, results in AUC equal to $66.37\%$. Similarly, observe in Table \ref{tab:gtsrb} that when GTSRB-CNN is under attack, VG achieves the best detection performance when it operated with brightness transformation and HSV equal to $200$ yielding an AUC equal to $97.81\%$.

\begin{figure}[t]
  \centering
    \includegraphics[width=1\linewidth]{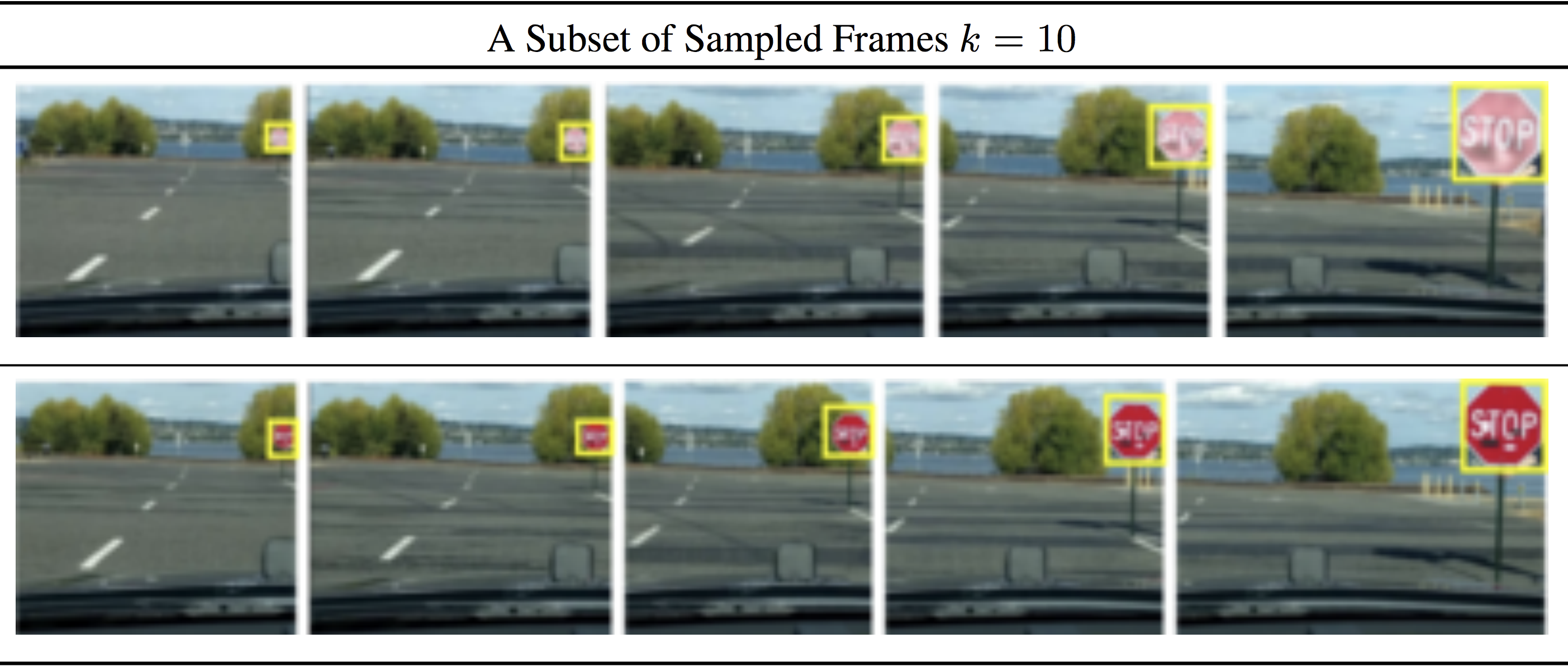}
  \caption{Drive-by experiment for LISA CNN conducted in \cite{eykholt2018robust}. The first and second row show stop signs manipulated by the poster-printed attack ($\ell_2$ loss function) and the sticker attack ($\ell_1$ loss function).}
  \label{fig:drivebytest}
\end{figure}

{\bf{Case Study: A Drive-by Experiment}}
In this section, we evaluate VisionGuard on the data collected in drive-by experiment conducted in \cite{eykholt2018robust}. Specifically, a smartphone camera is placed on a car, and obtain data at realistic driving speeds. The video is recorded at approximately 250 ft away from the sign while the driving track was straight without curves and the car speed varied between 0 mph and 20 mph. Recording was stopped once the vehicle passed the sign. Videos were recorded using clean and attacked signs. Images were extracted from the videos every $10$ frames on which classification is ran; see Figure \ref{fig:drivebytest} borrowed from \cite{eykholt2018robust}. The stop sign is attacked using (i) stickers - as discussed before - shown in the second row of Figure \ref{fig:drivebytest} and (ii) a poster-printing attack generated using the $\ell_2$ loss function, shown in the first row of Figure \ref{fig:drivebytest}. Note that the poster printed attack has been generated as in \cite{kurakin2016physical}. The major difference is that in the $\text{RP}_2$ attack, the perturbations are restricted to the surface area of the sign excluding the background while being robust to large angle and distance variations. In total, $33$ clean and $42$ adversarial (using both stickers and poster-printed attacks) images were generated. The accuracy of LISA CNN on the collected clean and adversarial images is $100\%$ and $4.76\%$, respectively.

\begin{figure}[t]
  \centering
   \subfigure[]{
      \label{fig:ROC1}
    \includegraphics[width=0.8\linewidth]{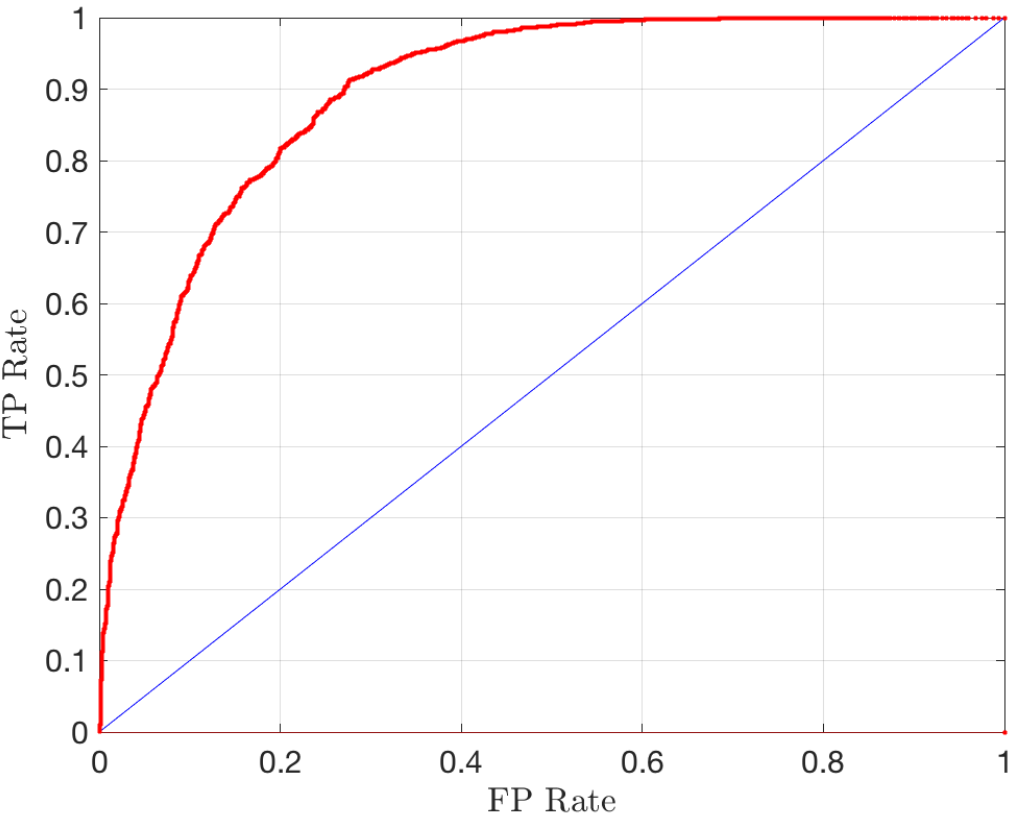}}
     \subfigure[]{
    \label{fig:ROC2}
     \includegraphics[width=0.8\linewidth]{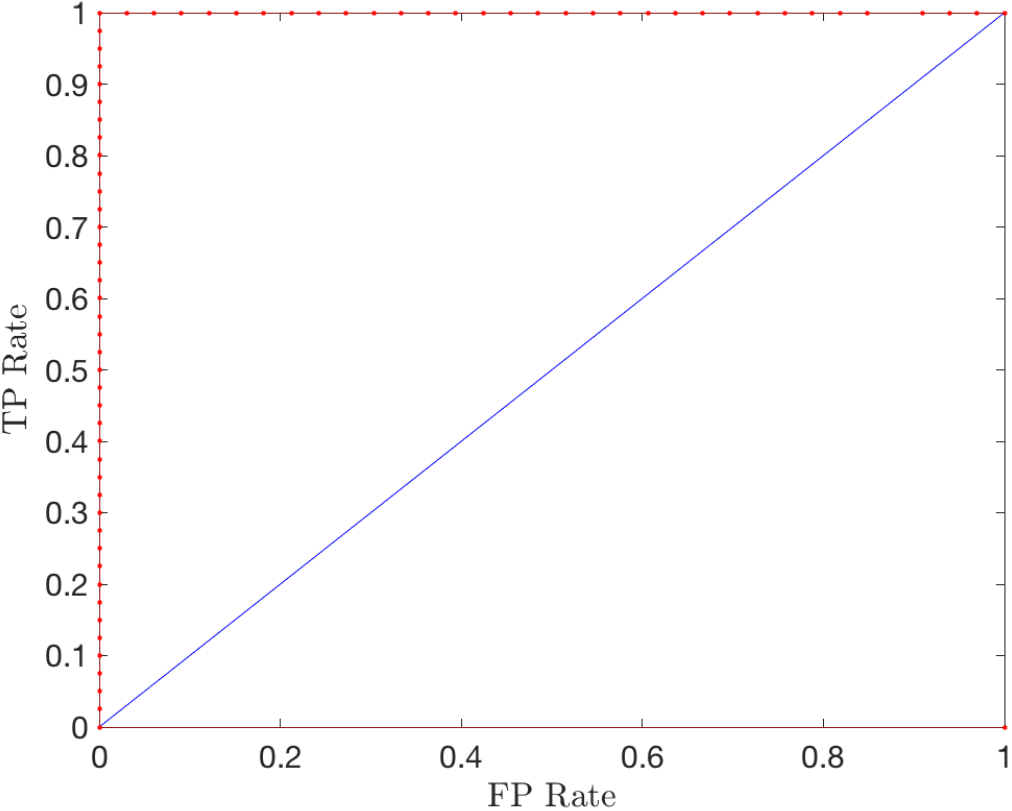}}
  \caption{ROC curve (red) corresponding to VisionGuard equipped with brightness transformation with $\text{HSV}=200$ on AdvNet (Figure \ref{fig:ROC1}) and the drive-by experiment (Figure \ref{fig:ROC2}); $\text{AUROC}=89.43\%$ for LISA CNN \cite{eykholt2018robust}. The detection threshold $\tau=0.43$ yields $\text{TP rate}=80\%$ and $\text{FP rate}=19\%$ on AdvNet and $\text{TP rate}=100\%$ and $\text{FP rate}=0\%$ on the drive-by experiment. Each point in the ROC curve corresponds to a detection threshold while the bblue line represents the ROC of a random detector.}

\end{figure}

The ROC curve for this small test dataset, which is not included in AdvNet, using VisionGuard with brightness transformation and $\text{HSV}=200$ is shown in Figure \ref{fig:ROC2}; observe that $AUC=100\%$. In fact, in this test dataset for a detection threshold $\tau=0.43$ VisionGuard yields TP and FP rate equal to $100\%$ and $0\%$, respectively. Recall that the same threshold on AdvNet yields TP and FP rate equal to $80\%$ and $19\%$, respectively.

\begin{table*}[]
\caption{Accuracy of CNNs on the Proposed Traffic Sign Dataset}.
\label{tab:accRP2}
\centering
\begin{tabular}{|l|l|l|l|l|l|l|l|l|l|}
\hline
Accuracy & \multicolumn{2}{l|}{\begin{tabular}[c]{@{}l@{}}Clean \\ Data\end{tabular}} & \begin{tabular}[c]{@{}l@{}}Adversarial \\ Data\end{tabular} & \begin{tabular}[c]{@{}l@{}}Clean \\ Stop\end{tabular} & \begin{tabular}[c]{@{}l@{}}Adversarial\\ Stop\end{tabular} & \begin{tabular}[c]{@{}l@{}}Clean Speed\\ Limit 35\end{tabular} & \begin{tabular}[c]{@{}l@{}}Adversarial\\ Speed Limit 35\end{tabular} & \begin{tabular}[c]{@{}l@{}}Clean \\ Yield\end{tabular} & \begin{tabular}[c]{@{}l@{}}Adversarial \\ Yield\end{tabular} \\ \hline
LISA     & \multicolumn{2}{l|}{82.72\%}                                               & 31.62\%                                                     & 78.14\%                                               & 36.03\%                                                    & 72.52\%                                                        & 18.51\%                                                              & 96.12\%                                                & 40.41\%                                                      \\ \hline
GTSRB    & \multicolumn{2}{l|}{-}                                                     & -                                                           & 82.24\%                                               & 21.33\%                                                    & -                                                              & -                                                                    & -                                                      & -                                                            \\ \hline
\end{tabular}
\end{table*}

\begin{table*}[]
\caption{LISA-CNN: Comparative Experiments in terms of AUC against KDE on AdvNet.}
\label{tab:LISAall}
\centering
\begin{tabular}{|l|l|l|l|l|l|l|l|l|l|l|l|l|l|l|}
\hline
LISA & B90     & B110    & B150    & B170    & {\bf{B200}}  & R5      & R15     & R20     & JPEG92 & Z3 & Z25 & Z28 & Z30 & Z31  \\ \hline
VG      & 79.95\% & 83.21\%  & 87.94\% & 89.00\% & \bf{89.43}\% & 78.18\% & 75.13\% & 71.20\% & 66.37\% & 27.13\% & 67.89\% & 51.32\% & 56.14\% & 60.84\% \\ \hline
KDE & \multicolumn{14}{l|}{~~~~~~~~~~~~~~~~~~~~~~~~~~~~~~~~~~~~~~~~~~~~~~~~~~~~~~~~~~~~~~~~~~~~~~~~~~~~~~~~~~~~~~~~~~~~~~~~~~~~~~~~~~~~~~~~~~81.56\%}   \\ \hline  
\end{tabular}
\end{table*}

\begin{table*}[]
\centering
\caption{GTSRB-CNN: Experiments in terms of AUC on AdvNet Stop Signs.}
\label{tab:gtsrb}
\begin{tabular}{|l|l|l|l|l|l|l|l|l|l|l|l|l|l|l|}
\hline
\begin{tabular}[c]{@{}l@{}}GTSRB\end{tabular} & B90     & B110    & B150    & B170    & {\bf{B200}}    & R5      & R15     & R20     & JPEG92 & Z3 & Z25 & Z28 & Z30 & Z31 \\ \hline
VG                                                            & 93.93\% & 95.15\% & 96.97\% & 97.43\% & \bf{97.81}\% & 90.76\% & 94.00\% & 92.61\% & 86.21\% & 3.36\% & 32.24\% & 33.34\% & 67.41\% & 83.11\% \\ \hline
\end{tabular}
\end{table*}

\begin{table*}[]
\label{tab:stopD}
\caption{LISA-CNN:  AUC-based performance of VG on the drive-by experiment.}
\centering
\begin{tabular}{|l|l|l|l|l|l|l|l|l|l|l|}
\hline
\begin{tabular}[c]{@{}l@{}}LISA\end{tabular} & B90 & B110 & B130 & B150 & B170 & {\bf{B200}} & R5 & R15 & R20 & JPEG92  \\ \hline
VG                                                            & 100\%      & 100\%       & 100\%       & 100\%       & 100\%       & \textbf{100}\%       & 91.89\%   & 99.92\%    & 100\%    & 77.35\% \\ \hline
KDE & \multicolumn{10}{l|}{~~~~~~~~~~~~~~~~~~~~~~~~~~~~~~~~~~~~~~~~~~~~~~~~~~~~~~~~~~~~~~~~~~~~~~~~~~~~~~~~~~~~~~~~~~~~~~~~~~~~~~~~~~~~~~~~~~99.63\%}   \\ \hline  
\end{tabular}
\end{table*}

\begin{table}[t]
\caption{Disk-space requirements of VisionGuard, MagNet, and KDE per dataset}
\centering
\label{tab:diskspace}
\begin{tabular}{|l|l|l|l|l|}
\hline
     & MNIST   & CIFAR10 & ImageNet & AdvNet \\ \hline
VG & 0 & 0  & 0 & 0       \\ \hline
MagNet  & 24KB   & 16KB    & DNF ($>2$ weeks) & X      \\ \hline
KDE & 5.1MB    & 4.2MB   & $\approx$ 4.8GB & 901KB \\ \hline
\end{tabular}
\end{table}

\begin{table}[t]
\caption{Average Execution Time (secs) per Image and Detector}
\label{tab:runtimeDet}
\begin{tabular}{|l|l|l|l|l|}
\hline
          & \begin{tabular}[c]{@{}l@{}}MNIST\\ (JPEG92)\end{tabular} & \begin{tabular}[c]{@{}l@{}}CIFAR10\\ (JPEG92)\end{tabular} & \begin{tabular}[c]{@{}l@{}}ImageNet\\ (JPEG92)\end{tabular} & \begin{tabular}[c]{@{}l@{}}AdvNet\\ (B200)\end{tabular} \\ \hline
VG & 0.0016                                                   & 0.0187                                                     & 0.08                                                        & 0.0179                                                   \\ \hline
MagNet  & 0.002   & 0.011    & X & X      \\ \hline
KDE  & 0.0006   & 0.0005    & 0.0003 & 0.0004\\ \hline
\end{tabular}
\end{table}

\subsection{Real-Time Computational Requirements}\label{sec:realtime}
In this section, we evaluate the real-time computational requirements of VG in terms of required disk-space and execution time. Specifically, in Table \ref{tab:diskspace}, we report the disk-space requirements for VG, MagNet, and KDE. MagNet requires storing the auto-encoders that are used at runtime. The auto-encoders used in \cite{meng2017magnet} are stored as keras models and require $24$ KB and $16$ KB for MNIST and CIFAR10, respectively. KDE requires disk-space to store the last hidden layer output of the DNN for all training images. In particular, KDE requires $5.1$ MB, $4.2$ MB, $4.8$ GB, and 901 KB  for MNIST, CIFAR10, ImageNet, and AdvNet images, respectively. 
In contrast, VG does not have any disk-space requirements, as it does not rely on training sets or building new DNNs and performs the image transformations in memory. In Table \ref{tab:runtimeDet}, we also report the runtime requirement of VG; note that these runtimes can be improved depending on the implementation but they are significantly smaller than the runtime required to craft adversarial attacks in Table \ref{tab:runtimeAtt}. 
Specifically, VG requires $0.0016$, $0.0187$, $0.08$, and $0.0179$ secs on average to check if an MNIST, CIFAR10, ImageNet, and AdvNet image is adversarial, respectively. 
Observe that as the dimensions of the input images increase, the runtime of VG increases slightly as well. This is expected as the JPEG compression/decompression time complexity is $O(n)$, where $n$ is the number of pixels \cite{chiou2017complexity}. Observe in the same table that MagNet and VG have similar execution times due to their similar structure but the KDE detector is faster than both as it does not require applying transformation to images. Nevertheless, as discussed earlier, its fast execution time is accompanied with poor performance on large-scale image domains; see Table \ref{tab:imagenet}. Finally, note that VG is fast enough for real time applications, such as autonomous cars. For instance, the YOLO neural network \cite{redmon2016you} typically operates in autonomous driving applications at 55 frames per second (FPS) (or at 155 FPS for a small accuracy trade off) i.e., $55$ images are generated per second \cite{yurtsever2020survey}. If VG requires $0.08$ secs to reason about trustworthiness of an $224\times224\times3$ image, then it can be used to analyze  $0.08 \text{secs}* 55 \text{FPS}=4.4$ frames per second. In other words, it can be used to analyze every $12$-th frame assuming that the car camera operates at $55$ FPS.  

\section{Conclusions}
In this paper, we proposed VisionGuard an attack- and dataset-agnostic defense against digital and physical adversarial input images to perception systems that scales to large-scale image domains. 
To determine whether an image is adversarial or not, VisionGuard checks if the output of the classifier remains consistent under label-invariant transformations that tend to squeeze out adversarial components. Finally, we also proposed AdvNet, the first dataset that includes clean and physically attacked images. 

\section{Acknowledgments}

This work was supported in part by AFRL and DARPA FA8750-18-C-0090, ARO W911NF-20-1-0080, ONR N00014-17-1-2012, and SRC  Task 2894.001. 


\bibliography{YK_bib}
\bibliographystyle{IEEEtran}



\end{document}